%% file: main.tex
\documentclass{article}
\usepackage{spconf,amsmath,graphicx}
\usepackage{amssymb}
\usepackage{subfigure}
\usepackage{algorithm}
\usepackage{algpseudocode}
\usepackage{url}
\usepackage{mathtools}
\usepackage{color}

\graphicspath{{./figures/}{./figures/draw/}}

\def\x{{\mathbf x}}

\def\I{{\mathbf 1}}

\def\f{{\mathbf{F}}}
\def\B{{\mathbf B}}
\def\b{{\mathbf b}}
\def\X{{\mathbf X}}

\def\W{{\mathbf W}}

\def\I{{\mathbf I}}
\def\1{{\mathbf 1}}
\def\S{{\mathbf S}}

\def\v{{\mathbf v}}
\newcommand\norm[1]{\left\lVert#1\right\rVert}
\DeclarePairedDelimiter{\ceil}{\lceil}{\rceil}

\title{Supervised Hashing with End-to-End Binary Deep Neural Network}
\name{Dang-Khoa Le Tan$^{\star}$ \qquad Thanh-Toan Do$^{\dagger}$ \qquad Ngai-Man Cheung$^{\star}$}
\address{$^{\star}$Singapore University of Technology and Design (SUTD) \\
    $^{\dagger}$The University of Adelaide}
    
\begin{document}
%
\maketitle
\input{abstract}
%
\input{intro}
\input{method}
\input{experiments}

\input{conclusion}



\section*{Acknowledgements}

This research is supported by the National Research Foundation Singapore under its AI Singapore Programme (Award number:
AISG-100E-2018-005)


\bibliographystyle{IEEEbib}
\bibliography{hash}

\end{document}

%% file: abstract.tex
\begin{abstract}
Image hashing is a popular technique applied to large scale content-based
visual retrieval due to its compact and efficient binary codes. Our work
proposes a new end-to-end deep network architecture for supervised hashing
which directly learns binary codes from input images and maintains hashing
properties, namely similarity preservation, independence, and
balancing.  Furthermore, we also propose a new learning scheme that copes
with the binary constrained loss function. The proposed algorithm not only is
scalable for learning over large-scale datasets but also outperforms
state-of-the-art supervised hashing methods, which are illustrated throughout
extensive experiments from various image retrieval benchmarks.
\end{abstract}


%% file: intro.tex
\section{Introduction and related works}
\label{sec:intro}

Content-based image retrieval (CBIR) is one of the interesting problems in
computer vision and has enormous applications such as being a common approach for
image captioning, visual searching (Google Images, Flickr).
State-of-the-art image search systems involve a major component which maps an
input image into a visual representation.
The output vector is searched through a database via Euclidean
distance-based comparison or quantization schemes in order to retrieve relevant
images~\cite{do2017embedding,tolias2016rmac}.
Due to the exponential growth of image data, it is necessary to reduce memory
storage and searching time in CBIR systems. An interesting approach for
achieving these requirements is binary hashing~\cite{deepbit, fasttraining,
feng2017deep, sah, do2016binary, toanlagrangian}.  Technically, instead of
producing a real-valued vector as a final representation, the hashing approach maps
an input into a compact
binary code via data-dependent or data-independent algorithms. Consequently,
the produced binary codes dramatically reduce memory storage. In addition, thanks
to the advance of computer hardware, we are able to compute the Hamming
distance between the binary codes within only one cycle clock via the
\texttt{POPCNT} instruction~\cite{wang2016learning}.

Data-dependent approaches utilize training datasets to learn
hashing models, and thus they usually outperform data-independent approaches.
Along with the spectacular rise of deep learning, recent data-dependent hashing methods
tried to construct end-to-end models which are able to simultaneously learn image
representations and binary codes~\cite{deepbit, fasttraining, feng2017deep,
do2016binary}. Thanks to joint optimization, the binary codes are able to
retain label information and so increasing the discrimination.
However, because the hashing network has to produce binary
output, the loss function involves binary constraints which are commonly
represented by a non-differentiable function, e.g., $sign$
\cite{bitscalable,Liong_2015_CVPR}.  One of the workarounds is to use
approximated functions of $sign$. For instance,~\cite{lai2015simultaneous} used a
\textit{logistic} function to relax the binary
constraints to range constraints. Although the proposed functions are
differentiable, these functions cause the vanishing
gradient problem when training via stochastic gradient descent (SGD)~\cite{hinton2006fast}. On
the other hand, \cite{liu2016deep} resolved the binary constraints by assuming
that the absolute function and $l_1$ regularization are differentiable
everywhere. However, it could be some degradation in performance due to the assumption.


In \cite{do2016binary}, the authors proposed a supervised hashing neural network (SH-BDNN). They used the idea of penalty
method~\cite{numericaloptim} to deal with the binary constraints on output codes and optimized the model via L-BFGS\@. Nevertheless, SH-BDNN is not an end-to-end model in which the feature representation and the binary code learning and not joint optimized. 




Our specific contributions are:
(i) We propose an end-to-end deep neural network (SH-E2E) for supervised
hashing which integrates 3 components of a visual search system, i.e., the
feature extraction component, the dimension reduction component, and the binary learning component.
(ii) We also introduce a
learning scheme which not only is able to cope with binary constraints but also
is scalable for large-scale training.
(iii) Solid experiments on 3 image retrieval benchmarks show
our method significantly outperforms other supervised hashing methods.

The remaining of this paper is organized as follows. Section~\ref{sec:method} presents the
proposed end-to-end hashing network and the learning scheme.
Section~\ref{sec:exp} describes the experiments and analyzes the results.
Section~\ref{sec:cons} concludes the paper.

%% file: method.tex
\section{Methodology}
\label{sec:method}

\subsection{Binary constrained loss function}
\label{ssec:binloss}


Given the set of $n$ images $\X = \{x_i\}_{i=1}^n$ and network parameters $\W$,
let $\f = f(\W,\X) \in \{-1, 1\}^{L \times n}$ be the output of the network, i.e., $\f$ is $n$ binary codes of length $L$ corresponding to the input.
We not only want the output to be binary but also to achieve other hashing properties. One of them is similarity preserving, i.e., samples belonging to same class should have similar codes, while samples belonging to different classes should have different codes. 
In other words, we want to minimize $\norm{{\frac{1}{L}} \f^T\f - \S}^2$  where:

\begin{equation}
\label{eq:S}
\S_{ij} = \left\{ \begin{array}{ll}
            -1 & \textrm{if $x_i$ and $x_j$ are not the same class} \\
            1  & \textrm{if $x_i$ and $x_j$ are the same class}
        \end{array} \right.
\end{equation}
From the definition mentioned, we define the loss function:

\begin{equation}
\begin{gathered}
    \min_{\W} 
        {\frac{\alpha}{2}}\norm{\frac{1}{L} \f^T\f - \S}^2
    \text{s.t}\; \f \in \{-1, 1\}^{L\times n}
\label{eq:nonrelax_loss}
\end{gathered}
\end{equation}
where hyper-parameter $\alpha$ controls the similarity property. Because of
the binary constraints under Equation~\ref{eq:nonrelax_loss}, the loss function becomes
a mixed-integer programming problem which is NP-hard. Inspired from~\cite{do2016binary},
we relax the constraints by introducing an auxiliary variable.
Let $\B \in \{-1, 1\}^{L \times n}$, the loss function is formulated as:
\begin{equation}
\begin{gathered}
    \min_{\W, \B} 
        \frac{\alpha}{2}\norm{\frac{1}{L} \f^T\f - \S}^2
       +  \frac{\beta}{2}\norm{\f-\B}^2 \\
    \text{s.t}\; \B \in \{-1, 1\}^{L\times n}
    \label{eq:relaxed_loss}
\end{gathered}
\end{equation}
The second term plays a role as a measure of constraint violation. If we set $\beta$
sufficiently, we can strongly force the output to be binary, which helps us easy
to reach the feasible solution. Additionally, by introducing $\B$, we can
optimize Equation~\ref{eq:relaxed_loss} by alternating methods
(Section~\ref{ssec:e2etraining}). Finally, we integrate the independence and balance
properties which were introduced in~\cite{Liong_2015_CVPR} by directly attaching them into the objective function:

\begin{equation}
\begin{gathered}
    \min_{\W, \B} 
        \frac{\alpha}{2}\norm{\frac{1}{L} \f^T\f - \S}^2
       + \frac{\beta}{2}\norm{\f-\B}^2 \\
       + \frac{\theta}{2}\norm{\f\f^T-\I}^2
       + \frac{\gamma}{2}\norm{\f\1_{n \times 1}}^2 \\
    \text{s.t}\; \B \in \{-1, 1\}^{L\times n}
\label{eq:loss}
\end{gathered}
\end{equation}

Where $\theta$, $\gamma$ are hyper-parameters; $\1_{n \times 1}$ is a $n$-dimensional column vector of ones.




\subsection{Network architecture}
\label{ssec:netarch}

\begin{figure}[t]
    \centering
    \includegraphics[scale=0.32]{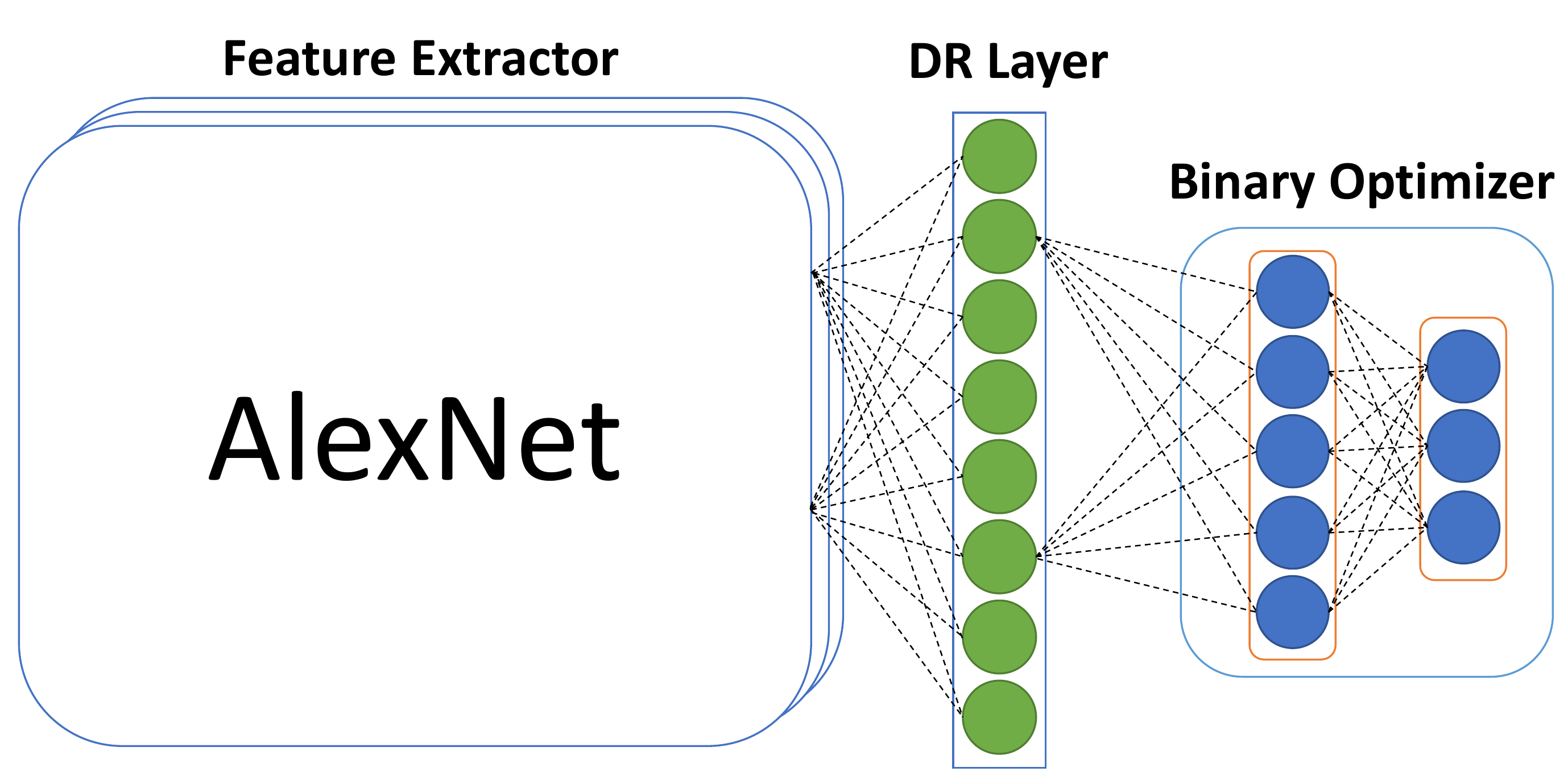}
    \caption{Overview of the proposed end-to-end hashing framework.}
\label{fig:e2enet}
\end{figure}

Figure~\ref{fig:e2enet} illustrates the overall architecture of the proposed end-to-end deep hashing network. The network is composed of three major components:
(i) a learning feature component which extracts visual image representations,
(ii) a dimension reduction layer, and
(iii) the hashing component which produces binary codes.

In CBIR systems, the feature extractor can be constructed from hand-crafted
features~\cite{sah} or learnable features such as convolutional neural
networks~\cite{tolias2016rmac}. In order to build an end-to-end system which can optimize all components, we choose the convolutional deep networks as the feature extractor. 
It is flexible that we can choose any convolutional deep networks such as
VGG~\cite{Simonyan14c}, AlexNet~\cite{krizhevsky2012imagenet}.
To make fair experiments, we utilize AlexNet as the
feature extractor of the network for which is consistent with other
hashing methods compared.  In the  proposed design, we remove the last layer (i.e.
\textit{softmax} layer) of AlexNet, and consider  its last fully connected
layer ($fc7$) as the image representation. Thus, the output of the feature
extractor component is a 1024-dimensional real-valued vector.


The dimension reduction component (the DR layer) involves a fully connected layer initialized
by PCA from the outputs of AlexNet's $fc7$ layer from the training set.
Specifically, let $\W_{PCA}$ and ${\b_{PCA}}$ be the weights and bias of the layer,
respectively:

\begin{equation}
\begin{aligned}
\W_{PCA} = [\v_1^T \; \v_2^T \; \cdots \; \v_p^T] \\
\b_{PCA} = - \W_{PCA} \mathbf{m}
\end{aligned}
\end{equation}

where $\v_1, \v_2, \ldots , \v_p$ are eigenvectors extracted from the covariance matrix and $\mathbf{m}$ is the mean of the $fc7$ features of the training set $\X$. We use the identity function as the activation function of this layer.


The last component, namely the hashing optimizer, is constructed from several
fully connected layers. The output of this component has the same length as
the length of required binary output.





\subsection{Training}
\label{ssec:e2etraining}

\begin{algorithm}[!t]
	\scriptsize
	\caption{End-to-End Supervised Hashing Deep Network Learning}
	\begin{algorithmic}[1]
		\Require
			\Statex $\X =\{\x_i\}_{i=1}^n$: training images;
            $L$: code length;
            $K$, $T$: maximum iteration;
            $m$: number of samples in a mini-batch.
		\Ensure
			\Statex
                Network parameters $\W$
			\Statex
            \State Initialize the network $\W_{(0)}^{(0)}$ via the pretrained AlexNet
            \State Initialize $\B^{(0)} \in \{-1,1\}^{L\times n}$ via ITQ~\cite{itq}
			\For{$k = 1 \to K$}
                \For {$t = 1 \to T $}
                    \State A minibatch $X_{(t)}$ is sampled from $\X$
                    \State Compute the corresponding similarity matrix $\S_{(t)}$
                    \State From $\B^{(k-1)}$, sample $B_{(t)}$ corresponding to $X_{(t)}$
                    \State Fix $B_{(t)}$, optimize $\W_{(t)}^{(k)}$ via SGD
                \EndFor
                \State Update $\B^{(k)} = \text{sign}(f(\W_{(T)}^{(k)}, \X)) $ 
            \EndFor

			\State Return $\W_{(T)}^{(K)}$
    \end{algorithmic}
\label{e2ealg}
\end{algorithm}

The training procedure is demonstrated in Algorithm~\ref{e2ealg} in which $\B^{(k)}$ is the binary codes of the training set at $k$th iteration and $\W_{(t)}^{(k)}$ is SH-E2E's parameters at the $t$th inner-loop of the $k$th outer-loop. First of all, we use a pretrained AlexNet as the initial weights of SH-E2E, namely $\W_{(0)}^{(0)}$. The DR layer is generated as discussed in Section~\ref{ssec:netarch} and the remaining layers are initialized randomly. In order to encourage the algorithm to converge faster, the binary variable $\B^{(0)}$ is initialized by ITQ~\cite{itq}. 

We apply the alternating approach to minimize the loss function (\ref{eq:loss}).
In each iteration $t$, we only sample a minibatch $X_{(t)}$ including $m$ images from the training set as well as corresponding binary codes $B_{(t)} $ from $\B^{(k)}$.  Additionally, we create the similarity matrix $\S_{(t)}$ (Equation~\ref{eq:S}) corresponding to $X_{(t)}$  as well as the $B_{(t)}$ matrix. Since $B_{(t)}\in \{-1, 1\}^{L \times m}$ has been already computed, we can fix that variable and 
update the network parameter $\W_{(t)}^{(k)}$ by standard back-propagation via SGD.
Through $T$ iterations, we are able to exhaustively use all training data.
After learning SH-E2E from the whole
training set, we update $\B^{(k)} = \text{sign}(f(\W_{(T)}^{(k)}, \X))$ and then start the
optimization procedure again until it reaches a criterion, i.e., after $K$ iterations.


Comparing to the recent supervised hashing SH-BDNN~\cite{do2016binary}, the proposed framework has two  advantages.
SH-E2E is an end-to-end hashing network which consists  both feature extraction and binary code optimization in a unified framework. It differs from SH-BDNN which requires image features as inputs. 
Secondly, thanks to the SGD, SH-E2E can be trained with large scale datasets because the network builds the similarity matrix from an image batch of $m$ samples ($ m \ll n$). It differs from SH-BDNN which uses the whole training set in each iteration.

%% file: experiments.tex
\section{Experiments}

\label{sec:exp}

\begin{figure*}[htb]
\centering
\subfigure[MNIST]{
    \includegraphics[scale=0.33]{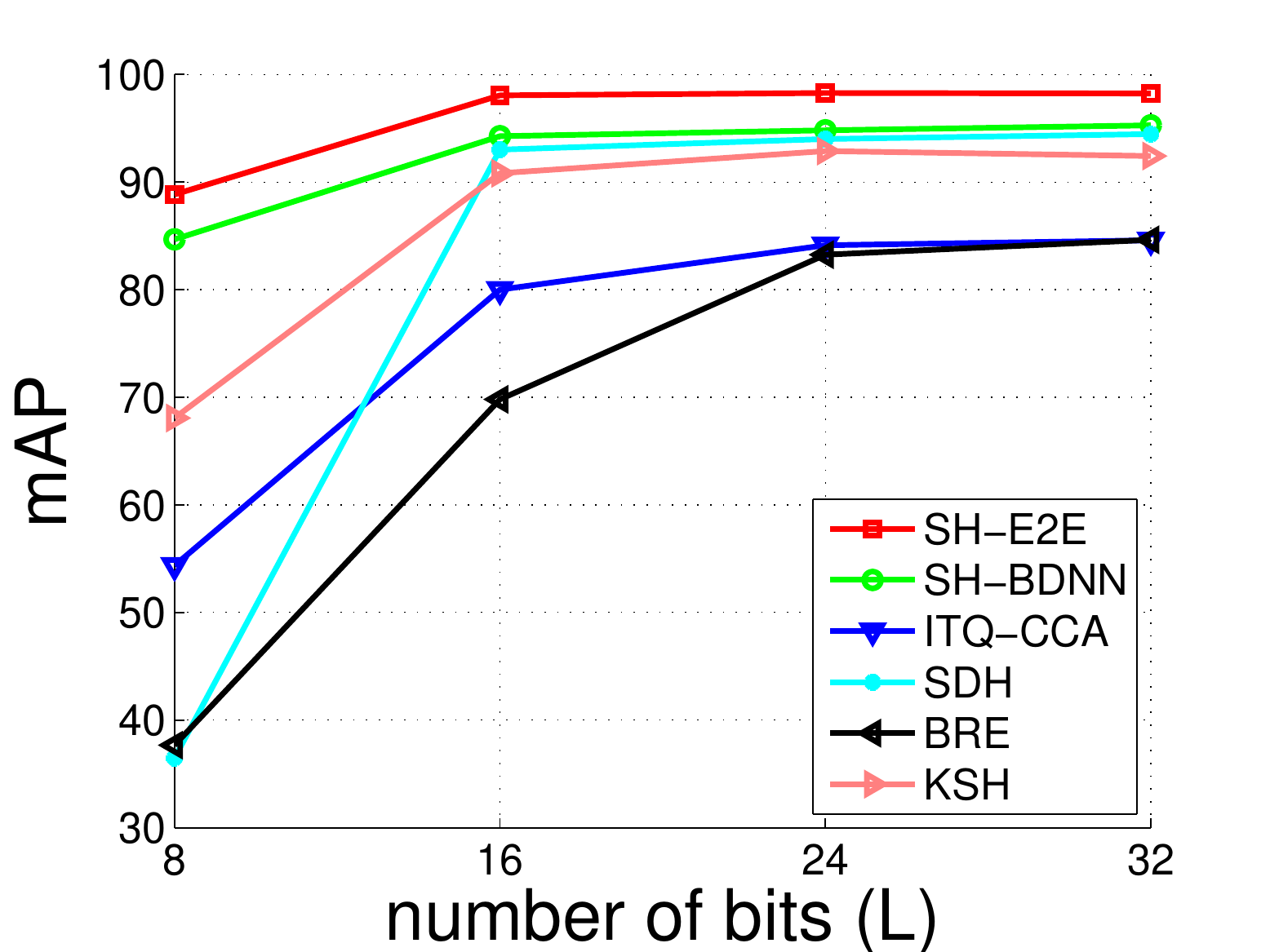}
    \label{fig:mnist}
}
\subfigure[Cifar10]{
    \includegraphics[scale=0.33]{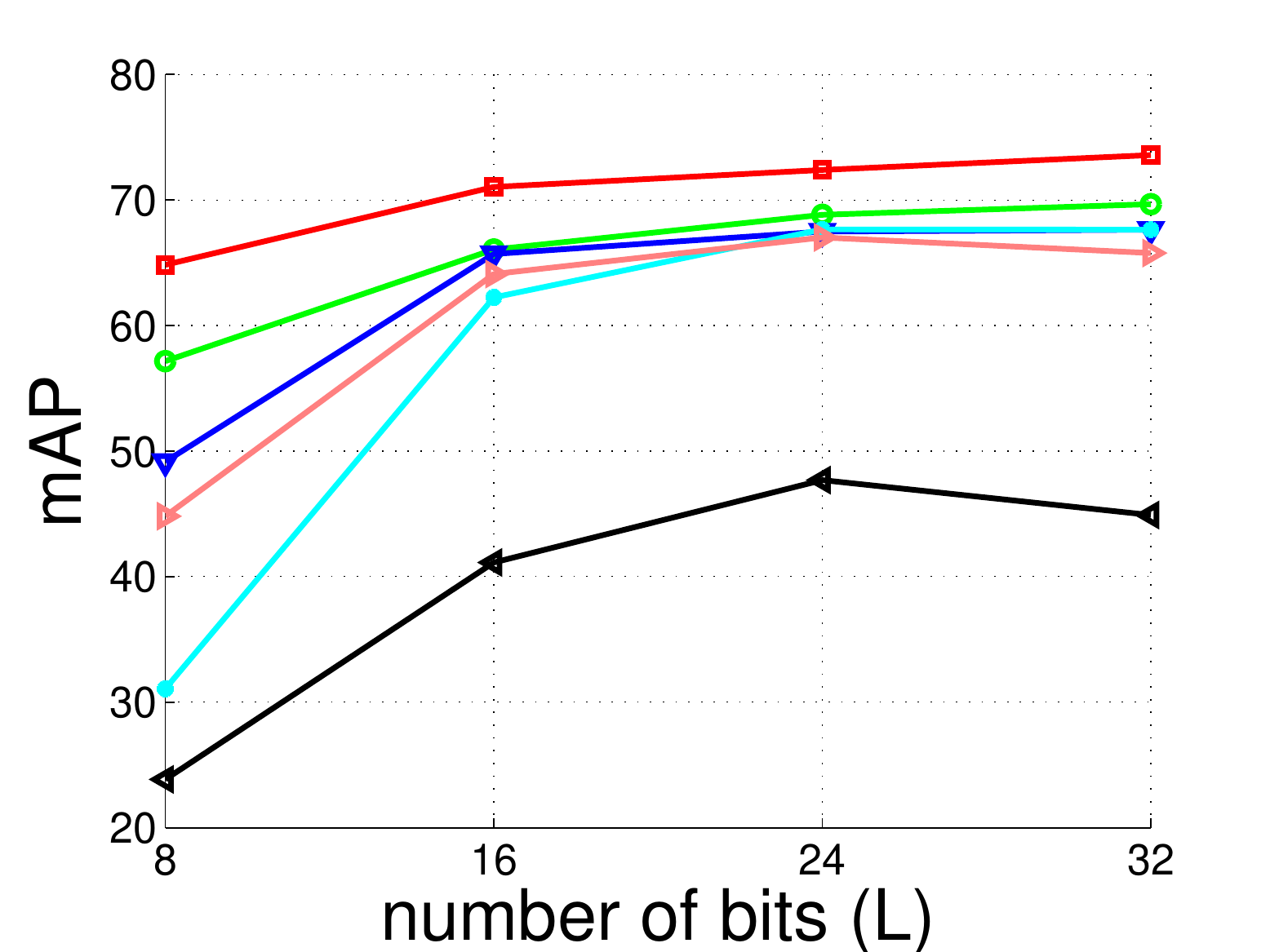}
    \label{fig:cifar10}
}
\subfigure[SUN397]{
    \includegraphics[scale=0.33]{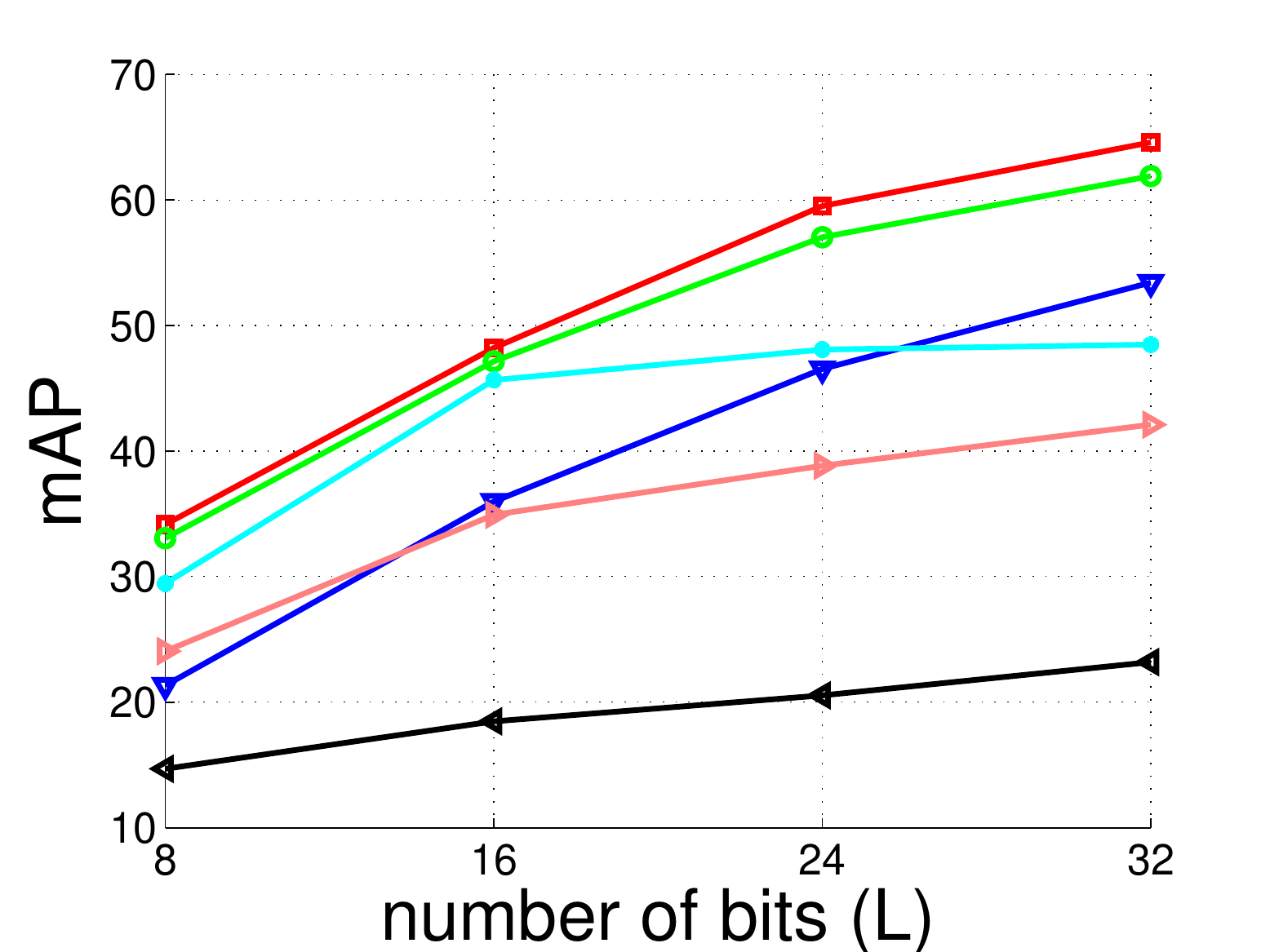}
    \label{fig:sun397}
}
\caption[]{mAP comparison between SH-E2E and other state-of-the-art methods
    on MNIST, Cifar10, SUN397.}
\label{fig:mnist_cifar_sun397}
\end{figure*}


\subsection{Datasets}
\label{ssec:datasets}

\textbf{MNIST}~\cite{mnist} comprises 70k grayscale images of 10 hand-written digits which are divided into 2 sets: 60k training images and 10k testing images. \\
\textbf{Cifar10}~\cite{cifar10} includes 60k RGB images categorized into 10 classes where images are divided into 50k training images and 10k testing images. \\
\textbf{SUN397}~\cite{sun397} is a large-scale dataset which contains 108754
images categorized into 397 classes. Following the setting
from~\cite{toanlagrangian}, we select 42 categories which have more than 500
images. This results 35k images. For each class,  we randomly sample 100 images as test samples and therefore we get 4200 images to form a testing set. The remaining images are considered as training samples.


\subsection{Implementation details}
\label{ssec:impl}

We implement SH-E2E by MATLAB and the MatConvNet
library~\cite{matconvnet}. All experiments are
conducted in a workstation machine (Intel(R) Xeon(R) CPU E5-2650 \@ 2.20GHz) with one Titan X GPU.
The last component includes 3 fully connected layer in which the sigmoid function is
used as the activation function. The number of units in the binary optimizer
are empirically selected as described in Table~\ref{tab:num_units}.

\begin{table}[t]
    \centering
    \caption{The number of units in the binary optimizer.}
    \begin{tabular}{|c|c|c|c|}
        \hline
        $L$      &Layer 1     &Layer 2    &Layer 3    \\ \hline \hline
        8      &90          & 20        & 8         \\ \hline
        16     &90          & 30        & 16        \\ \hline
        24     &100         & 40        & 24        \\ \hline
        32     &120         & 50        & 32        \\ \hline
        48     &140         & 80        & 48        \\
        \hline
    \end{tabular}
    \label{tab:num_units}
\end{table}

Regarding the hyperparameters of the loss function, we finetune
$\alpha$ in range $10^{-3}$ to $10^{-1}$, $\beta=10^{-2}$,
$\gamma=10^{-2}$ and $\theta=10^{-3}$. For training the network, we select
learning rate $lr=10^{-4}$ and weights decay to be $5\times 10^{-4}$. The size of a 
minibatch is $m=256$ for all experiments. Other settings in the algorithm are
set as following: $K=5$, $T=\ceil{\frac{4n}{m}}$.

\subsection{Comparison with other supervised hashing methods}

We compare the proposed method with other supervised hashing methods, i.e.,
SH-BDNN~\cite{do2016binary}, ITQ-CCA~\cite{itq},
KSH~\cite{liu2012supervised}, BRE~\cite{kulis2009learning}, SDH~\cite{Shen_2015_CVPR}.
In order to make fair comparisons for other methods, follow~\cite{do2016binary}, we use the pre-trained network AlexNet to
extract features from fully connected layer $fc7$ then use PCA to map
4096-dimensional features to lower dimensional space, i.e., a 800 dimensional space. The reduced features are used as inputs for compared methods.  

The comparative results between methods are shown in Fig.~\ref{fig:mnist_cifar_sun397}.  
On MNIST dataset, SH-E2E achieves fair improvements over compared methods. 
The performance of SH-E2E is saturated when the code lengths are higher than 16, i.e, its mAP are 98.03\%, 98.26\% and 98.21\% at code length 16, 24, 32, respectively.
The similar results can be observed on the Cifar10 dataset.  As shown in
Figure~\ref{fig:cifar10}, SH-E2E outperforms other supervised hashing methods
with a fair margin. SH-E2E outperforms the most competitive SH-BDNN from 4\% to 7.5\% at different code lengths. 
On the SUN397 dataset, SH-E2E and the second best SH-BDNN achieve comparable results, while these two methods significantly outperform other methods. 



\subsection{Comparison with other end-to-end hashing networks}
\label{ssec:compare_e2e}

We also compare the proposed deep network with other end-to-end supervised hashing
architectures, i.e., Lai et al.~\cite{lai2015simultaneous}, DHN~\cite{deephashingnet},
DQN~\cite{dqn}, DSRH~\cite{dsrh}, and DRSCH~\cite{bitscalable}. 
For end-to-end networks comparisons, we follow two different settings for the MNIST and the Cifar10 datasets. 
\begin{itemize}
    \item Setting 1: according to~\cite{dqn, deephashingnet,lai2015simultaneous}, we randomly select 100 samples per class to
        form 1k testing images. The rest 59k images are database images. The
        training set contains 5k images which are sampled from the database (i.e., 500 images per class).
    \item Setting 2: following~\cite{bitscalable, dsrh}, for test set, we sample 1k
        images for each class. This results 10k testing images. In the test
        phase, each test image is searched through the test set by the
        leave-one-out procedure. The rest 59k images are served as the training
        set.
\end{itemize}

The results of~\cite{lai2015simultaneous, deephashingnet,
bitscalable, dsrh, dqn}
are directly cited from the corresponding papers. Although those works proposed
approaches which simultaneously learn image features and binary codes by
combining CNN layers and a binary quantization layer, they used approximation
of $sign$ function, i.e.,~\cite{dsrh} and ~\cite{lai2015simultaneous} use
$logistic$ function, while~\cite{bitscalable} uses $tanh$ that may degrade the
performance.
Thanks to the new learning scheme for dealing with binary constraints
and the effectiveness of proposed network architecture, the proposed SH-E2E outperforms other
end-to-end methods with fair margins as observed in Table~\ref{tab:e2e_simul}
and Table~\ref{tab:e2e_bitscale}. Specifically, under Setting 1, the proposed SH-E2E and DHN~\cite{deephashingnet} achieve comparable results while these two methods significantly outperform other methods. Under Setting 2, the proposed SH-E2E consistently outperforms the compared DRSCH~\cite{bitscalable}  and DSRH~\cite{dsrh}   at all code lengths. 


\begin{table}[!t]
   \centering
   \caption{mAP comparison between SH-E2E, DHN~\cite{deephashingnet}, DQN~\cite{dqn}, and Lai et al.\cite{lai2015simultaneous}
       on Cifar10 (Setting 1).}
\begin{tabular}{|c|c|c|c|}
\hline
$L$                                         & 24 &32 &48 \\ \hline
SH-E2E                                    & 60.02 &61.35 &63.59 \\ \hline
DHN~\cite{deephashingnet}                 & 59.40 &60.30 &62.10 \\ \hline
Lai et al.\cite{lai2015simultaneous}             & 56.60 &55.80 &58.10 \\ \hline
DQN~\cite{dqn}                            & 55.80 &56.40 &58.00 \\ \hline
\end{tabular}
\label{tab:e2e_simul}
\end{table}

\begin{table}[!t]
   \centering
   \caption{mAP comparison between SH-E2E, DRSCH\cite{bitscalable}
       and DSRH~\cite{dsrh}
       on Cifar10 (Setting 2).}
\begin{tabular}{|c|c|c|c|}
\hline
$L$                       &24 &32 &48 \\ \hline
SH-E2E                    &67.16 &68.72 &69.23 \\ \hline
DRSCH~\cite{bitscalable}  &62.19 &62.87 &63.05 \\ \hline
DSRH~\cite{dsrh}          &61.08 &61.74 &61.77 \\ \hline
\end{tabular}
\label{tab:e2e_bitscale}
\end{table}


%% file: conclusion.tex
\section{Conclusions}
\label{sec:cons}

We propose a new deep network architecture which efficiently
learns compact binary codes of images. The proposed network comprises three components,
i.e., feature extractor, dimension reduction and binary code optimizer. These components are trained in an end-to-end framework. 
In addition,
we also propose the new learning scheme which can cope with binary constraints and also allows the network to be trained with large scale datasets. 
The experimental results on three benchmarks show the improvements of the proposed
method over the state-of-the-art supervised hashing methods.